# Practical Comparable Data Collection for Low-Resource Languages via Images

**Aman Madaan, Shruti Rijhwani, Antonios Anastasopoulos,**
**Yiming Yang & Graham Neubig**
Language Technologies Institute
Carnegie Mellon University
Pittsburgh, PA 15213, USA
{amadaan,srijhwan,aanastas,yiming,gneubig}@cs.cmu.edu

## Abstract

We propose a method of curating high-quality comparable training data for low-resource languages with monolingual annotators. Our method involves using a carefully selected set of images as a pivot between the source and target languages by getting captions for such images in both languages independently. Human evaluations on the English-Hindi comparable corpora created with our method show that 81.1% of the pairs are acceptable translations, and only 2.47% of the pairs are not translations at all. We further establish the potential of the dataset collected through our approach by experimenting on two downstream tasks – machine translation and dictionary extraction. All code and data are available at https://github.com/madaan/PML4DC-Comparable-Data-Collection

## 1 Introduction

Machine translation (MT) is a natural language processing task that aims to automatically translate text from a source language into a target language. Current state-of-the-art methods for MT are based on neural network architectures (Barrault et al., 2019), and often require large parallel corpora (i.e. the same text in two or more languages) for training.

Creating parallel data between a source and target language usually requires bilingual translators who are fluent in both languages. While there are a few language pairs for which translators are readily available, for many low-resource languages pairs it is challenging to find *any* speakers sufficiently proficient in both the source and target languages. We propose a method to create data for training machine translation systems in such situations. Our method relies on obtaining captions for images depicting concepts that are relevant in most languages in the world. The captions are collected individually in each of the source and target languages, removing the requirement for the human annotators to be well-versed in both languages. The captions for the same image are then paired to create training data. Unlike existing methods (Hewitt et al., 2018; Bergsma & Van Durme, 2011; Singhal et al., 2019; Hitschler et al., 2016; Chen et al., 2019) aimed at leveraging multimodal information for creating comparable corpus, our method does not rely on the existence of large resources available in both the languages. Our goal is to propose a solution for the *cold start* scenario: one in which there is absolutely no comparable data available for the source language.

Notably, as the captions are developed independently for each language, our method creates data that are *comparable*, rather than strictly parallel. Nevertheless, comparable corpora have been proven useful for machine translation and related applications (Munteanu et al., 2004; Abdul-Rauf & Schwenk, 2009; Irvine & Callison-Burch, 2013). In this work, we evaluate the utility of our collected data as a replacement for parallel corpora in two ways:

- We show that bilingual speakers (of the source and target languages) judge the dataset as containing over 81% acceptable translation pairs.

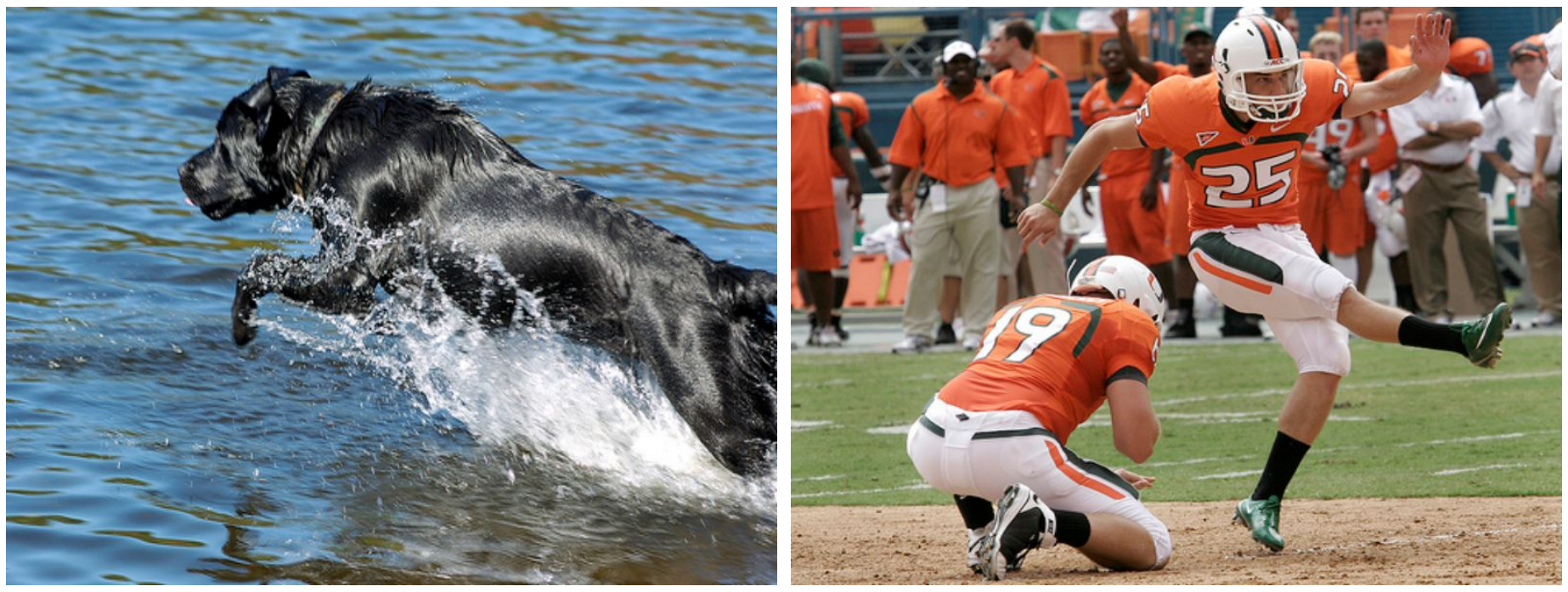

Figure 1: Images depicting a concise, universal concept (left) and a complex culture-specific activity (right). The images are taken from the Flickr8k dataset (Hodosh et al., 2013).

- We demonstrate that the data collected via our method has the potential of being used for downstream tasks such as machine translation and dictionary extraction.

Moreover, we also compare our method to the traditional process of parallel corpus creation and show that it is significantly more cost-efficient.

We apply our proposed method and evaluation techniques to a specific language pair: Hindi as the source language and English the target. Hindi is chosen as a testbed because, although it has over 520 million speakers (Chandramouli & General, 2011), far fewer parallel corpora exist as compared to other widely spoken languages like French, German, Spanish, etc. (Kunchukuttan et al., 2017). We note that our annotation tools and methodology are not language-specific and can be easily adapted to any very low-resource setting with two assumptions: i) availability of speakers, and ii) a writing system. If the language uses a novel character set not present in unicode, we can create a mapping from existing unicode symbols to those in the language. If (digitally) literate speakers are not available, or if the language lacks a working orthography (most of the world's languages are indeed oral (Simons & Fennig, 2017)) we could instead collect spoken input, to be transcribed to a phonetic transcription, with tools such as Adams et al. (2019) or Li et al. (2020).

## 2 Methodology

**Overview** Our method requires a set of $N$ images $\mathbb{I} = \{\mathrm{I}_1, \mathrm{I}_2, ..., \mathrm{I}_N\}$. Each image $\mathrm{I}_i$ is associated with a set of $P(\geq 1)$ captions $\mathbb{C}_i^{trg} = \{C_{i,1}^{trg}, C_{i,2}^{trg}, ..., C_{i,P}^{trg}\}$ in the target language $trg$. The $P$ captions are provided by different annotators. Note that the captions are only required in the (presumably high-resource) target language.

For each image $i$, $Q(\geq 1)$ annotators are then asked to provide captions in the source language $src$, yielding $\mathbb{C}_i^{src} = \{C_{i,1}^{src}, C_{i,2}^{src}, ..., C_{i,Q}^{src}\}$. We experiment with two different ways of obtaining comparable data $\mathbb{D}_{en-fr}$ from the sets $\mathbb{C}_i^{src}$ and $\mathbb{C}_i^{trg}$. In the first method , we take Cartesian product of the set of captions in the two languages.

$$\mathbb{D}_{en-fr} = \bigcup_{i=1}^{N} \mathbb{C}_i^{trg} \bigotimes \mathbb{C}_i^{src} \tag{1}$$

$$\text{Where } \mathbb{C}_i^{trg} \bigotimes \mathbb{C}_i^{src} = \{(C_{i,j}^{trg}, C_{i,k}^{src}) : j \in [1, P], k \in [1, Q]\} \tag{2}$$

Thus, yielding $P * Q$ *comparable* sentences per image. A second way is to randomly assign each of the sentences in source set to the target set, yielding $min(P, Q)$ comparable sentences per image. We refer to these methods as *cross* and *random* assignment respectively.

**Selecting the Right Images** The most suitable images for our task are those that are **simple** and **universal**. Images showing a simple event or an action involving common entities will be good candidates for our task. On the other hand, images sourced from news articles typically involve a large amount of context and will not be a good fit (Hodosh et al., 2013). Similarly, images that show an entity that is only popular within a certain geographical or cultural context (like a celebrity) will also not be suitable.

To capture *simplicity*, we define the caption complexity of an image $i$ by measuring the lexical variance, $d_i$, in its captions ($\mathbb{C}_i^{trg}$). We define $d_i = l_i + w_i + e_i$ where $l_i$, $w_i$, and $e_i$ are respectively the total length, the total number of unique words, and the total pairwise edit distance for the captions included in $\mathbb{C}_i^{trg}$:

$$l_i = \sum_{j=1}^{P} length(C_{i,j}^{trg}) \tag{3}$$

$$w_i = \sum_{j=1}^{P} unique_words(C_{i,j}^{trg}) \tag{4}$$

$$e_i = \sum_{j=1}^{P} \sum_{k=j+1}^{P} edit_dist(C_{i,j}^{trg}, C_{i,k}^{trg}) \tag{5}$$

$$d_i = l_i + w_i + e_i. \tag{6}$$

In other words, the score helps in finding images whose captions are short, have fewer unique words, and are consistent across annotators.

For example, consider the images shown in Figure 1 taken from the Flickr8k dataset (Hodosh et al., 2013). The image on the right (with a high caption complexity score) depicts a complex activity and is bound to attract varied captions. This becomes clear from the captions for this image from the dataset: “A holder and kicker for a football team dressed in orange, white and black play while onlookers behind them watch.” and “Two young men on the same football team are wearing orange and white uniform and playing on an outside field while coach and other players watch.” On the other hand, the image on the left shows a simpler concept. The captions for the image are almost identical, and are simple variations of “A black dog is running in the water.” We focus on such images as they will likely get consistent captions in any language, leading to higher quality comparable data.

Capturing the concept of universality is non-trivial, but several heuristics can be used for selecting such images. For example, allowing only those images that show a certain fixed set of entities (like trees, the sun, etc.) would be one such heuristic. In this work, however, we do not explicitly fix the set of allowed entities and events, but instead start with a dataset containing relatively generic images (Hodosh et al., 2013) of everyday events and actions involving people and animals (see Section §3 for details). We then further run the above-stated algorithm to select relatively simple images, and finally prune the non-universal images manually according to the subjective judgement of an annotator.

## 3 Experiments and Results

For our experiments, we used the Flickr8k dataset (Hodosh et al., 2013) mentioned above. The Flickr8k dataset has 8092 images, and each image has $P = 5$ English captions. We selected this dataset for two reasons: i) it contains images that depict everyday actions and events involving people and animals (favors universal images), and ii) it aims to include images that can be unambiguously described in a sentence (favors simple images). We first selected 700 images with the lowest caption complexity score (see section 2). Then, we manually pruned the set of images to retain $N = 500$ images to maximize the number of different concepts covered by the images, and to remove any image depicting a non-universal concept.

We used Hindi as the source language. We asked five different crowd workers, sourced via Amazon Mechanical Turk, to describe each of the images in the source language. The crowd workers needed to be fluent *only* in the source language (Hindi) and no information in the

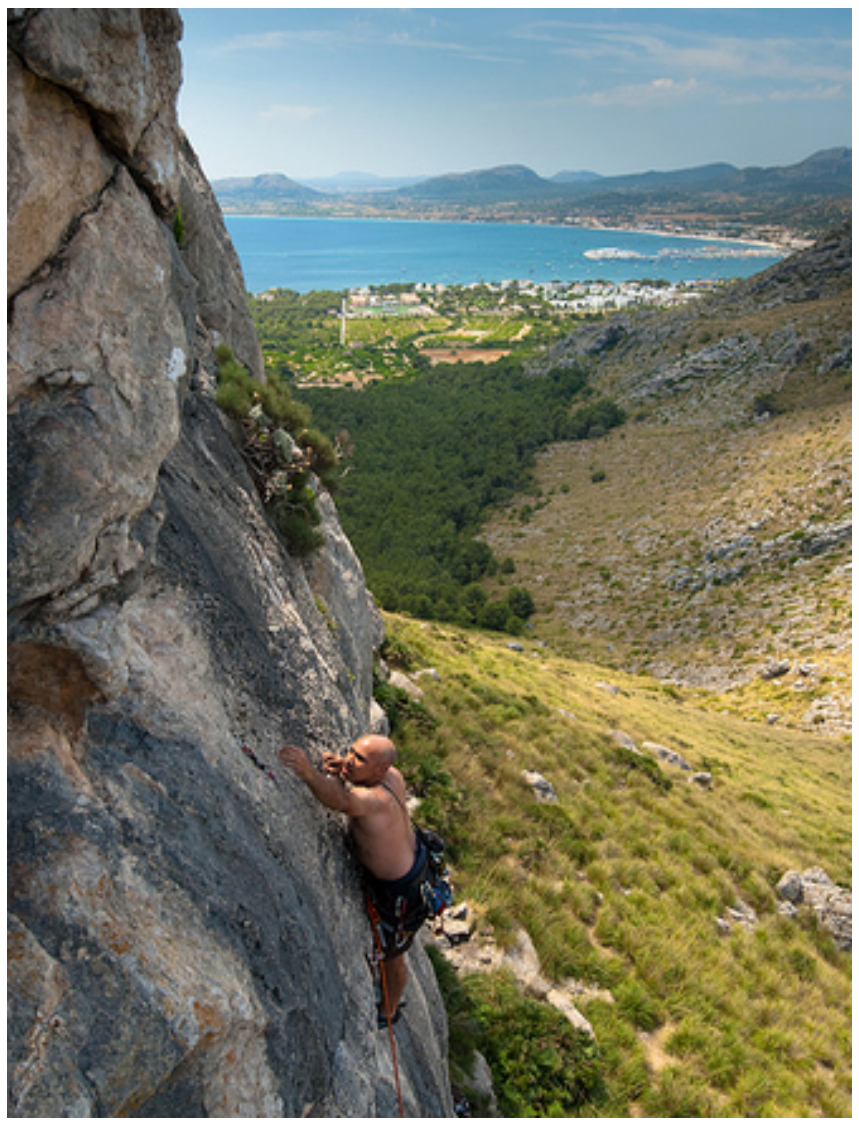

**English Captions from Flickr8k**
A bald , shirtless man rock climbing .
A bald man climbing rocks .
A man climbing up a rocky cliff
A man with no shirt on is rock climbing .
A rock climber scales a mountain .

**Crowdsourced Hindi Captions**
एक आदमी बिना शर्ट पहने चट्टान पर चड़ रहा है
कुछ लोग पहाड़ी पर चढ़ रहे हैं
एक आदमी पहाड़ पर चढ़ रहा है |
एक आदमी पहाड़ी पे ट्रैकिंग करता हुआ
एक आदमी पहाड़ पर चढ़ाई कर रहा है

**Translated Hindi Captions (for reference)**
A man is climbing a rock without wearing a shirt
Some people are climbing the hill
A man is climbing a mountain.
A man trekking up a hill
A man is climbing a mountain

Figure 2: An example image and the corresponding English (part of the dataset) and Hindi (generated by the workers using only the image) captions.

target language is supplied or requested. Only workers located in India were allowed to participate in the task.

Following Hodosh et al. (2013), we provided the annotators with guidelines (Table 3 in the appendix) that are conducive to reducing the variance in image captions. We obtained $Q = 5$ captions per image, for a total of 2500 captions for the 500 images. For quality control, the authors who are native Hindi speakers manually verified all the captions. Note that our method *requires no resources in the source language* apart from the instructions for the annotators. We make no assumptions specific to Hindi in our setup, and it can be adopted for any other language.

### 3.1 Evaluating Data Quality

**Manual Evaluation** Out of the 500 images selected for the experiment, we randomly chose 120 for a final human evaluation on the quality of the data obtained. We used comparable data obtained via the *cross* method (see Section 2). Therefore, a total of $120 * 25 = 3000$ sentence pairs were evaluated. The annotators were asked to rate the translation quality on a Likert scale (Likert, 1932) between 1-5. Table 1 shows the instructions for the evaluators and the % of translation falling under each quality category.

As Table 1 shows, over 81% of the sentences were cumulatively rated to be acceptable or better. More importantly, only 2.47% of sentences were rated as not being translations at all, and 16% were deemed perfect translations. For a low resource setting, this means that the data created with our proposed method can be used without any further pruning. An example image and the corresponding captions are shown in Figure 2, and additional examples can be found in the appendix.

### 3.2 Evaluating Quality of Comparable Data on Downstream Tasks

We present two different tasks to measure the quality of our data. We use the *random* variant of the dataset (2500 comparable sentence pairs) for both the experiments (Section 2). Given that the small number of data points, the goal of these experiments is to show that the dataset is comparable in nature and thus if replicated at a larger scale.

**Unsupervised Dictionary Extraction** The task of unsupervised dictionary extraction from parallel data aims at creating a table containing pairs of translated words and phrases. Our approach relies on word-level alignments between our collected captions. We first generate word-level alignments using `fast_align` (Dyer et al., 2013) to create word alignment data $A$. From $A$, we generate a dictionary of word pairs $\{(w_{\text{src}}, w_{\text{tgt}}) : \text{count}_{\text{align}}(w_{\text{src}}, w_{\text{tgt}}) > c \wedge P_{\text{align}}(w_{\text{tgt}}|w_{\text{src}}) > p\}$. Here $\text{count}_{\text{align}}$ is the number of times $w_{\text{tgt}}$ and $w_{\text{src}}$ are aligned in $A$, $P_{\text{align}}$ is the probability of alignment as observed in $A$, and $c$ and $p$ are two tied hyper-parameters. We use $(p, c) \in \{(0.5, 20), (0.6, 5), (0.9, 2)\}$, chosen by tuning on the development set. We manually evaluated the generated word pairs and found that out of 75 word pairs generated by our method, 43 (57.3%) were accurate. Some sample translation pairs induced from the dictionary are shown in the Appendix.

| **Quality** | **Criteria** | **%** | **Cum. %** |
|---|---|---|---|
| Perfect | The translation is flawless. | 16.45 | 16.45 |
| Good | The translation is good. The differences between a perfect and a good translation are not very important to the meaning of the source sentence | 35.22 | 51.67 |
| Acceptable | The translation conveys the meaning adequately but can be improved | 29.43 | 81.10 |
| Bad | The translation conveys the meaning to some degree but is a bad translation | 16.42 | 97.52 |
| Not a translation | There is no relation whatsoever between the source and the target sentence | 2.47 | 100.00 |

Table 1: Detailed instructions for the evaluators and quality evaluation results

**Translation Downstream Task** We use a Transformer based neural machine translation system for all translation experiments (Vaswani et al., 2017). Each transformer has 4 attention heads with a 512 dimensional embedding layer and hidden state. Dropout (Srivastava et al., 2014) with a 0.3 probability is used in all layers. We use BPE tokenization (Sennrich et al., 2015) with a vocabulary of size 8000 for all the experiments. Given that we had a small amount of data, the hyper-parameters were not tuned. We used the defaults that work well for other low-resource scenarios.

We use the TED Hindi-English parallel corpus (Kunchukuttan et al., 2017) in conjunction with our dataset (OURS) to measure the potential improvements when additionally using our collected data. The TED data has pre-defined training and test splits with 18798 and 1243 sentences, respectively. We create training and test splits for OURS with 2220 training and 248 test sentences. We train two models: i) $\text{MODEL}_{\text{TED + OURS}}$ trained on a combination of our data (OURS) and the ted data (TED), and ii) $\text{MODEL}_{\text{TED}}$ trained only on the TED data. We evaluate our models using BLEU (Papineni et al., 2002) scores on varying amounts of training data on the TED and OURS test set. BLEU is a standard metric used by the MT community which captures the syntactic similarity between the expected and the predicted translations. As the results in Table 2 show, for both the test sets, the gains from adding the OURS dataset are the highest when the TED model is trained with only 500 instances. For the TED test set, increasing the TED data further initially hurts the BLEU scores, possibly due to a domain mismatch. However, with enough TED data, OURS data still helps the BLEU scores. We also note that the BLEU scores are consistently higher on the OURS test set. While the number of data points (2500) is not enough for training and evaluating machine translation systems, these results from our proof of concept setting indicate that if the exercise is repeated at scale, the collected data could be useful for training machine translation systems.

### 3.3 Cost Breakdown

We used Amazon Mechanical Turk (MTurk) to obtain 2500 captions for the 500 images. Seventy-six workers participated in the job for a total cost of 197 USD. The average time required to caption each image was 4.04 minutes, i.e., a total of about 168 hours for 2500 captions. Note that as per the recommendation of the Government of India (GOI, 2019), the highest recommended minimum wage in India across all zones is 447 INR/day or 6.27 USD/day. Assuming an 8-hour working day, this is 78.37 cents per hour – we paid ≈50% higher, at 117.2 cents per hour (all workers were required to be located in India).

| | Eval: TED test | | Eval: OURS test | |
|---|---|---|---|---|
| #TED Samples | $\text{MODEL}_{\text{TED}}$ | $\text{MODEL}_{\text{TED + OURS}}$ | $\text{MODEL}_{\text{TED}}$ | $\text{MODEL}_{\text{TED + OURS}}$ |
| 500 | 0.07 | 0.36 | 0.07 | **5.26** |
| 1250 | 1.41 | 0.93 | 0.13 | **4.72** |
| 2500 | 2.33 | 1.49 | 0.17 | **4.18** |
| 5000 | 3.56 | 2.85 | 0.16 | **6.90** |
| 9000 | 6.00 | 6.33 | 0.56 | **5.26** |
| 18000 | 11.36 | 11.36 | 0.83 | **6.36** |

Table 2: Machine Translation BLEU Scores on the TED and OURS test set. The differences are *not* statistically significant for the TED test set (using the paired sample t-test, $p = 0.05$. We **highlight** statistically significant improvements).

On average, professional translators charge about 0.1 USD/word or 31.56 USD/hour for English-Hindi translation.[1] Given that the 2500 English captions had a total of 114,433 words, professional translation would have charged about 5,539 USD by hourly rates or 12,107 USD by the per-word rate. Thus, our method is **at least about 28 times cheaper even after paying higher than the market rate**. We would like to reiterate that our method is most suitable for situations where no bilingual speakers are available for the language pairs of interest. However, the cost analysis shows that our proposed method offers a cost-effective solution in scenarios where comparable data is required at scale.

## 4 Conclusion and Future Work

In this work, we propose a method that uses images for generating high-quality comparable training data without the need for bilingual translators. More specifically, our technique for image selection and crowdsourcing results in useful training data for scenarios where finding annotators proficient in both the languages is challenging, as demonstrated by human evaluation and downstream task performance.

To the best of our knowledge, we are the first to introduce the idea of crowdsourcing comparable data using images for low resource settings. The IAPR TC-12 dataset (Grubinger et al., 2006) is one of the earliest image captioning dataset that comprises of travel images with captions in English, Spanish, and German provided by tour guides. The datasets released by Elliott et al. (2016) (30,000 images with captions in German and English) and Funaki & Nakayama (2015) (1000 images with captions in English and Japanese) were both obtained with the help of professional translators. Hewitt et al. (2018) and Bergsma & Van Durme (2011) rely on a corpus of images associated with words (accessed via image search engines) in the languages of interest. Similarities in images are then used to induce bilingual lexicons. In contrast, our method is ideal for settings where absolutely no resources are available for a low resource language. Further, Singhal et al. (2019) use a similar proprietary dataset obtained via Google search to learn multilingual embeddings. Hitschler et al. (2016) and Chen et al. (2019) improve the quality of statistical machine translation and bilingual lexicon induction by using large monolingual image-captioning corpora. While their work is orthogonal to ours, it underscores the fact that the dataset generated by our method can indeed boost downstream tasks.

In the future, we plan to use our data creation technique on extremely low-resource languages and release parallel corpora that can potentially propel the use of state-of-the-art NLP techniques on these languages. It would also be interesting to explore methods to quantify the definition of universality and select such images for tasks like ours.

[1] `https://search.proz.com/employers/rates`

## Acknowledgments

The authors are thankful to the annotators for their work, and to the anonymous reviewers for their thoughtful comments. This material is based upon work generously supported partly by the National Science Foundation under grant 1761548. Shruti Rijhwani is supported by a Bloomberg Data Science Ph.D. Fellowship. This material is based on research sponsored in part by the Air Force Research Laboratory under agreement number FA8750-19-2-0200. The U.S. Government is authorized to reproduce and distribute reprints for Governmental purposes notwithstanding any copyright notation thereon. The views and conclusions contained herein are those of the authors and should not be interpreted as necessarily representing the official policies or endorsements, either expressed or implied, of the Air Force Research Laboratory or the U.S. Government. Thanks to Waqar Hasan for the helpful discussion.

# A Appendix

- Table 3 shows the instructions to the annotators, translated from the instructions used by (Hodosh et al., 2013).
- Figure 3 shows example of images with high and low caption diversity scores.
- Table 4 shows sample word pairs from the dictionary learned by our method. Figures 4-8 show some sample images and the corresponding English captions and the Hindi captions provided by the annotators.

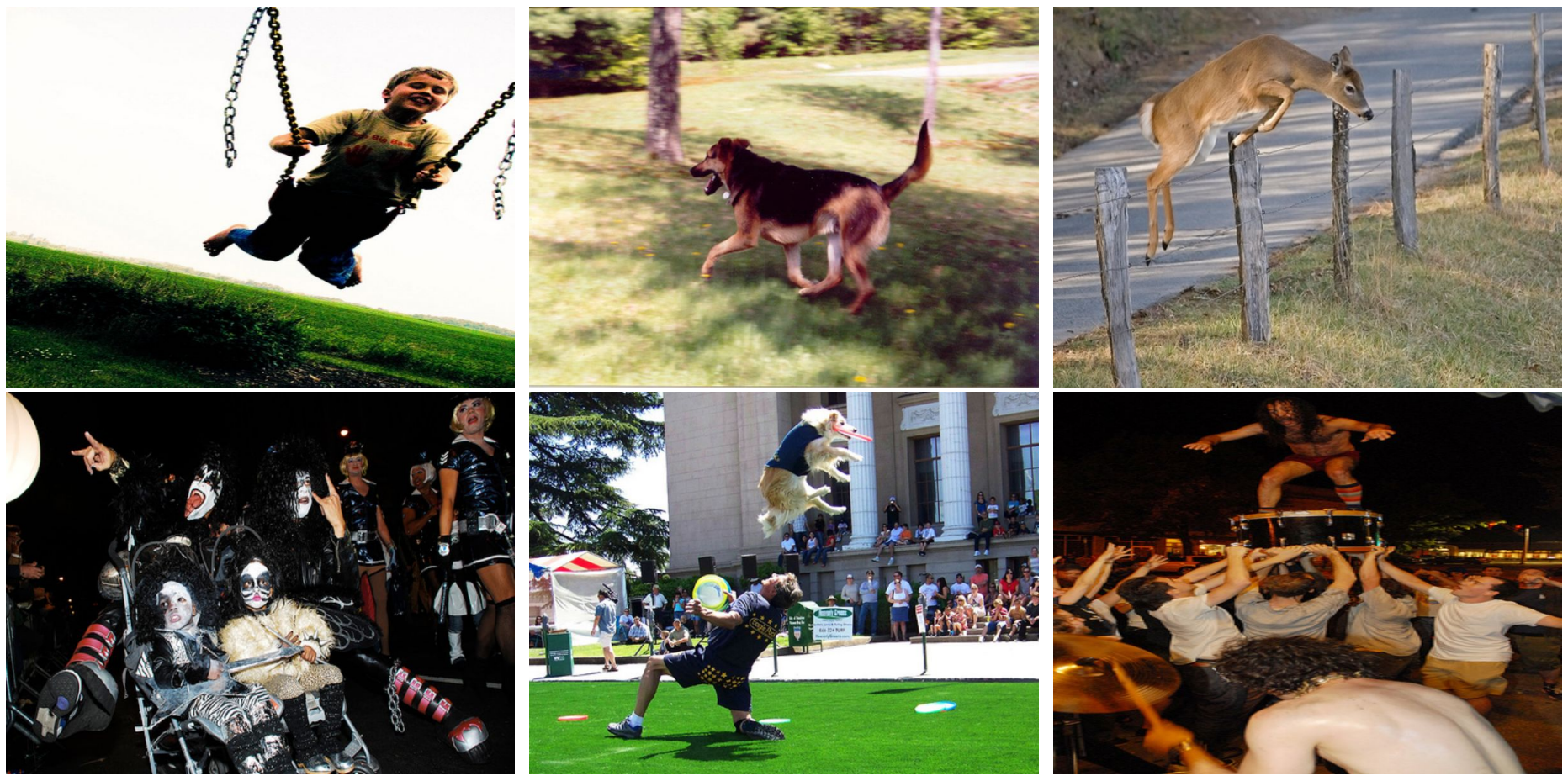

Figure 3: Some images with low (top row) and high (bottom row) diversity scores

| **Instructions (en)** | **Instructions (hi)** |
|---|---|
| 1. You must describe each image with one sentence.<br>2. Please provide an accurate description of the activities, people, animals and objects you see depicted in the image.<br>3. Each description must be a single sentence.<br>4. The description should be written using Hindi script.<br>5. Try to be concise.<br>6. If you don't know the meaning of a concept in Hindi, you can use English to express it.<br>7. Please pay attention to grammar and spelling.<br>8. You don't have to use perfect Hindi for this task. Describe the image as you see fit. | 1. आपको एक वाक्य के साथ हर छवि का वर्णन करना होगा।<br>2. कृपया उन गतिविधियों, लोगों, जानवरों और वस्तुओं का सटीक विवरण प्रदान करें जिन्हें आप चित्र में देख रहे हें ।<br>3. प्रत्येक विवरण एक ही वाक्य का होना चाहिए।<br>4. विवरण हिंदी में लिखा जाना चाहिए।<br>5. संक्षिप्त होने का प्रयास करें।<br>6. अगर किसी शब्द का हिंदी में मतलब ना पता हो तो उसे इंग्लिश में ही लिख दें ।<br>7. व्याकरण और वर्तनी पर ध्यान दें।<br>8. छवि के वर्णन में शुद्ध हिंदी का उपयोग करना ज़रूरी नहीं हे । जैसा ठीक लगे लिखें । |

Table 3: Instruction for the workers for the caption task. Note that only the Hindi instructions were supplied to the workers; the English translation is shown here as a reference.

| | | |
|---|---|---|
| हैट | hat | ✓ |
| लड़का | boy | ✓ |
| इस | a | ✗ |
| कुत्ते | dogs | ✓ |
| ट्रैम्पोलिन | trampoline | ✓ |
| कुत्ता | dog | ✓ |
| गुलाबी | girl | ✗ |
| चार | four | ✓ |
| स्नान | in | ✗ |
| किक | blue | ✗ |
| बच्चा | boy | ✓ |
| महिलायें | women | ✓ |
| लड़की | girl | ✓ |

Table 4: Randomly Sampled Correct (✓) and Incorrect (✗) Word Pairs from the Generated Dictionary

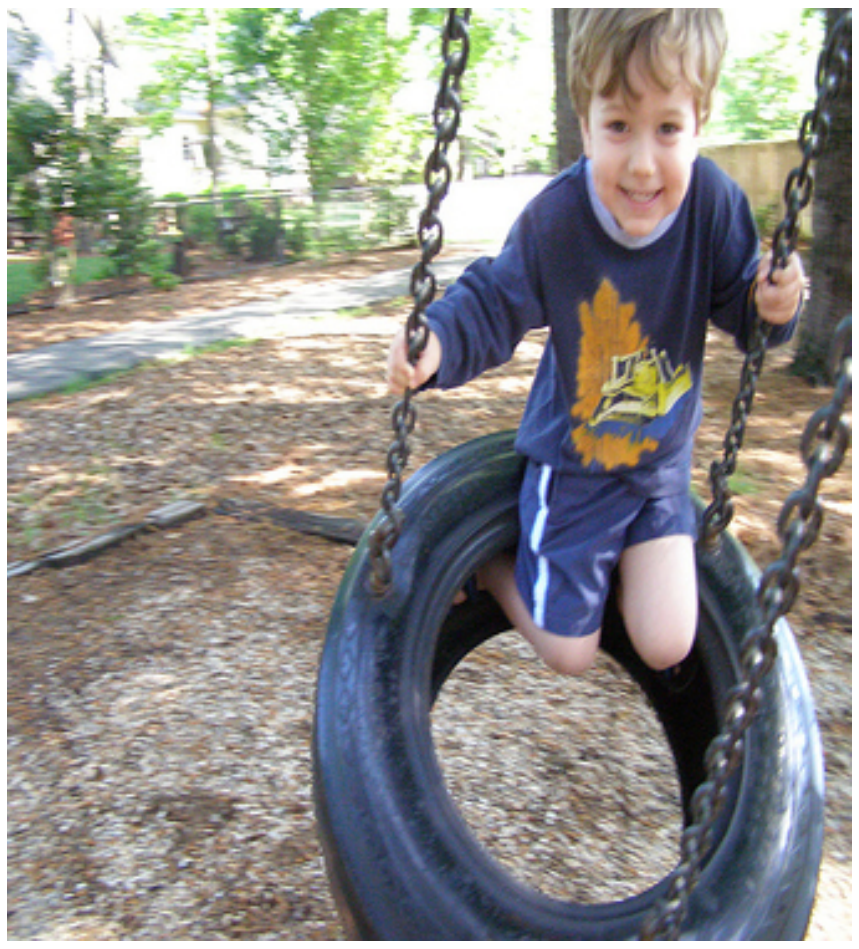

| |
|---|
| **Crowdsourced Hindi Captions** |
| एक बच्चा झूले पर खेल रहा है |
| एक बच्चा टायर पर बैठा है |
| बच्चा खेल रहा है |
| बच्चा झूला झूलता है |
| **English Captions from Flickr8k** |
| A boy rides on a tire swing . |
| A boy swinging on a tire-swing . |
| A small boy in a horizontal tire swing . |
| Young boy smiles at the camera from the tire swing . |
| Young boy swinging on a tire swing . |
| **Translated Hindi Captions (for reference)** |
| A child is playing on a swing |
| A child is sitting on a tire |
| a child is playing |
| A child is swinging on a swing |

Figure 4: An Example Image and the Corresponding English (part of the dataset) and Hindi (generated by the workers using only the image) captions.

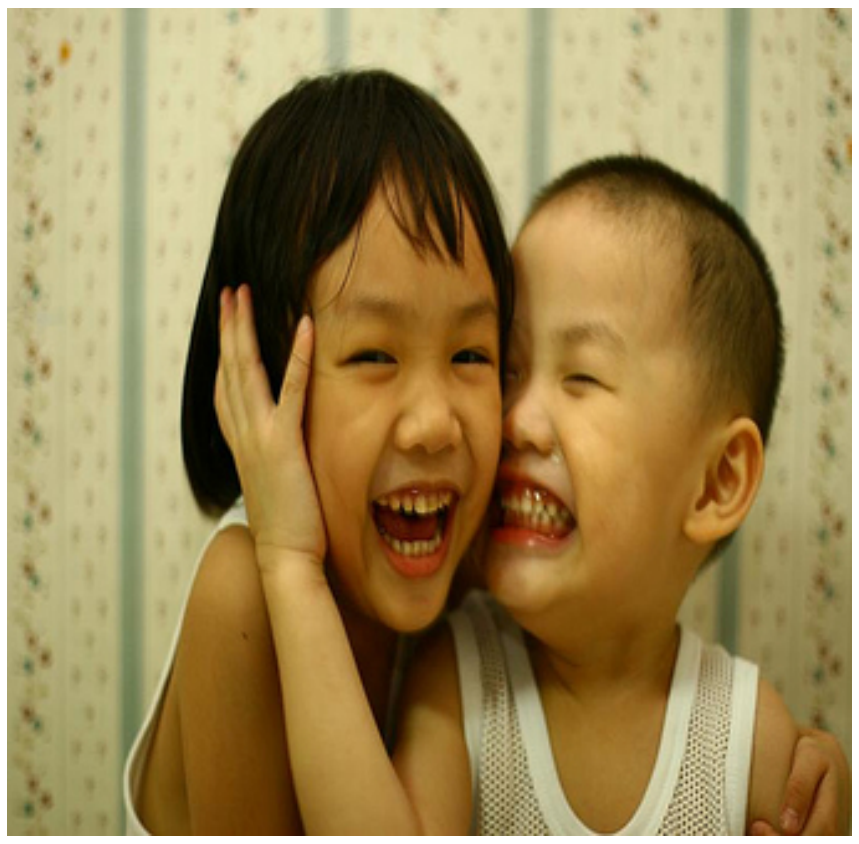

**Crowdsourced Hindi Captions**
दो बच्चे खेल रहे हैं
दो बच्चे आपस में गले लग कर हँस रहे हैं
दो बच्चे आपस में खेल रहे है
बच्चा खेल रहा है
बच्चा बच्ची को हग कर रहा है

**English Captions from Flickr8k**
Two asian children smile while hugging each other .
Two children grin and hug each other .
Two children hug and smile .
Two children smile for the camera .
Two kids hugging each other .

**Translated Hindi Captions (for reference)**
two children are playing
Two children are hugging each other and laughing
Two children are playing together
A child is playing
A child is hugging another child

Figure 5: An Example Image and the Corresponding English (part of the dataset) and Hindi (generated by the workers using only the image) captions.

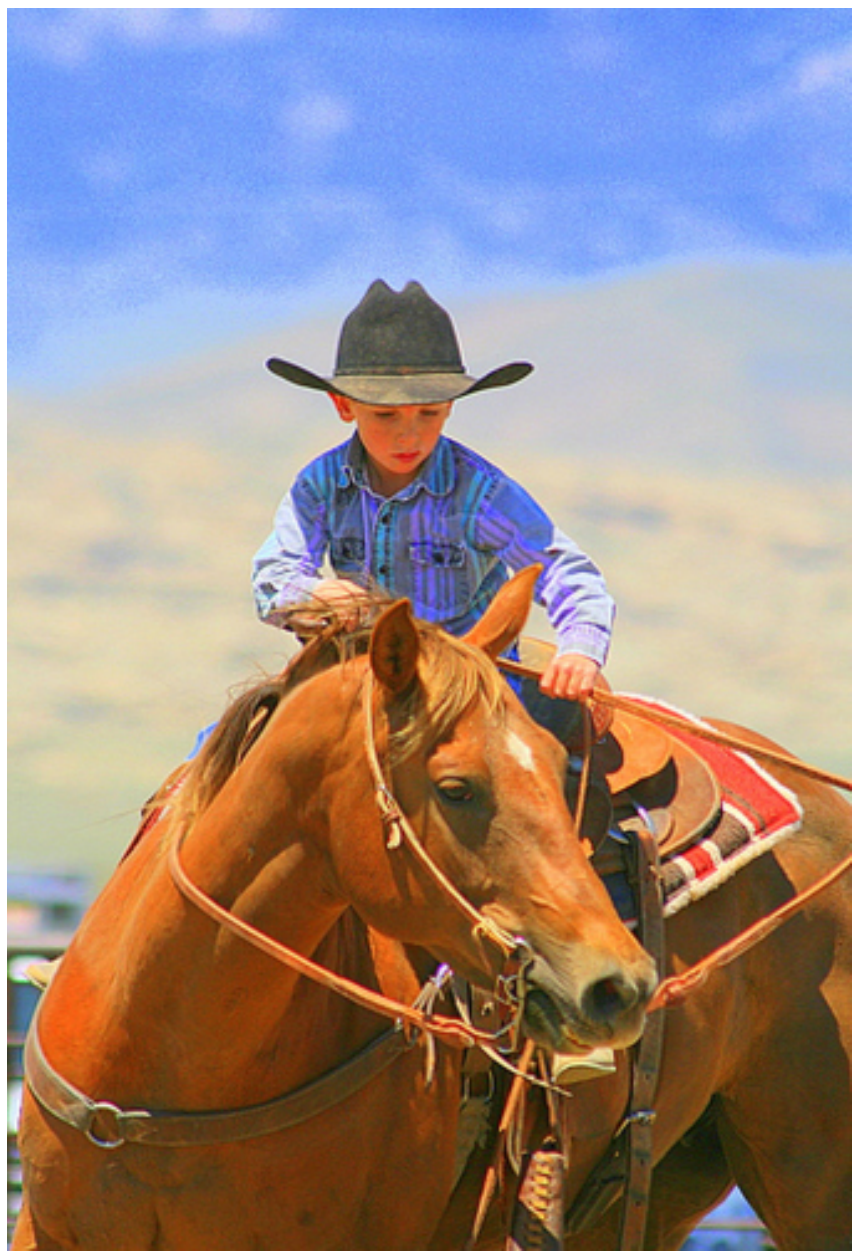

**Crowdsourced Hindi Captions**
एक छोटा बच्चा घोड़े के ऊपर बैठा हुआ है
एक छोटा बच्चा घोड़े पर बैठा है
एक बच्चा घोड़े पर बैठा है
लड़का हॉर्स राइड कर रहा है
घोड़े पर एक छोटा लड़का

**English Captions from Flickr8k**
A young boy in a cowboy hat rides a horse .
A young boy in a cowboy hat rides a large horse .
A young boy rides a red horse .
Boy riding a horse .
a young boy wearing a cowboy hat and riding on a horse

**Translated Hindi Captions (for reference)**
A small child is sitting on a horse
A small child is sitting on a horse
A child is sitting on a horse
A boy is riding a horse
A little boy on a horse

Figure 6: An Example Image and the Corresponding English (part of the dataset) and Hindi (generated by the workers using only the image) captions.

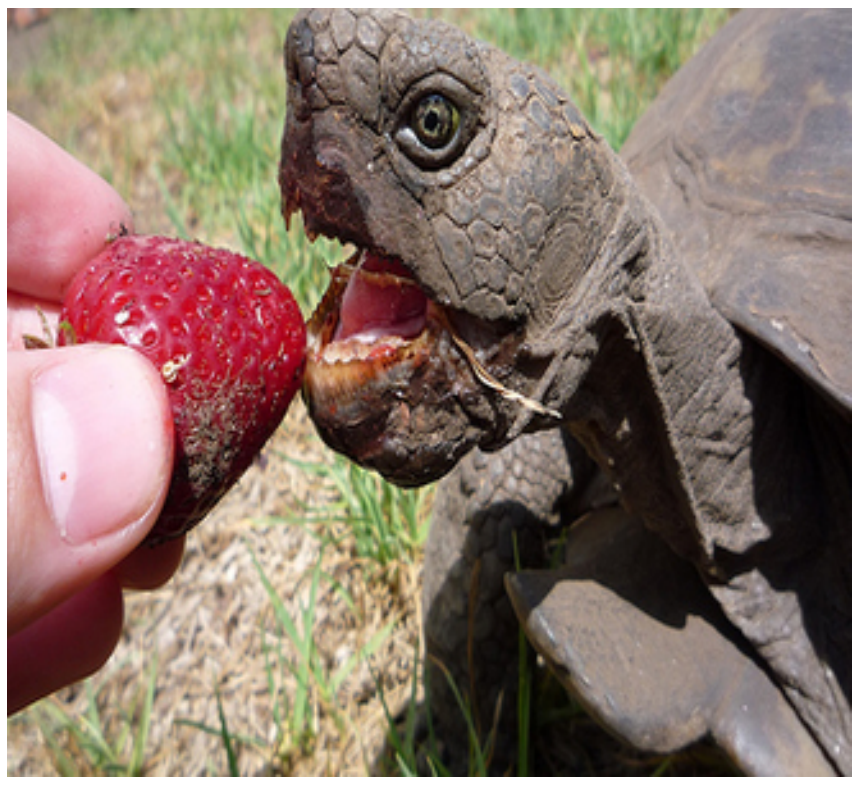

**Crowdsourced Hindi Captions**
एक आदमी एक कछुए को स्ट्रॉबेरी खिला रहा है
एक कछुआ स्ट्रॉबेरी खा रहा है
कछुवे को स्ट्रॉबेरी खिलाई जा रही है
कछुवा स्ट्रॉबेरी खा रहा है.

**English Captions from Flickr8k**
A person feeding a strawberry to a turtle
A tortoise attempts to eat a berry .
Someone feeds a strawberry to a turtle .
The tortoise is being fed an strawberry .
The turtle is being fed a strawberry

**Translated Hindi Captions (for reference)**
A man is feeding strawberries to a turtle
A turtle is eating strawberries
Strawberries are being fed to the tortoise
The tortoise is eating strawberries.

Figure 7: An Example Image and the Corresponding English (part of the dataset) and Hindi (generated by the workers using only the image) captions.

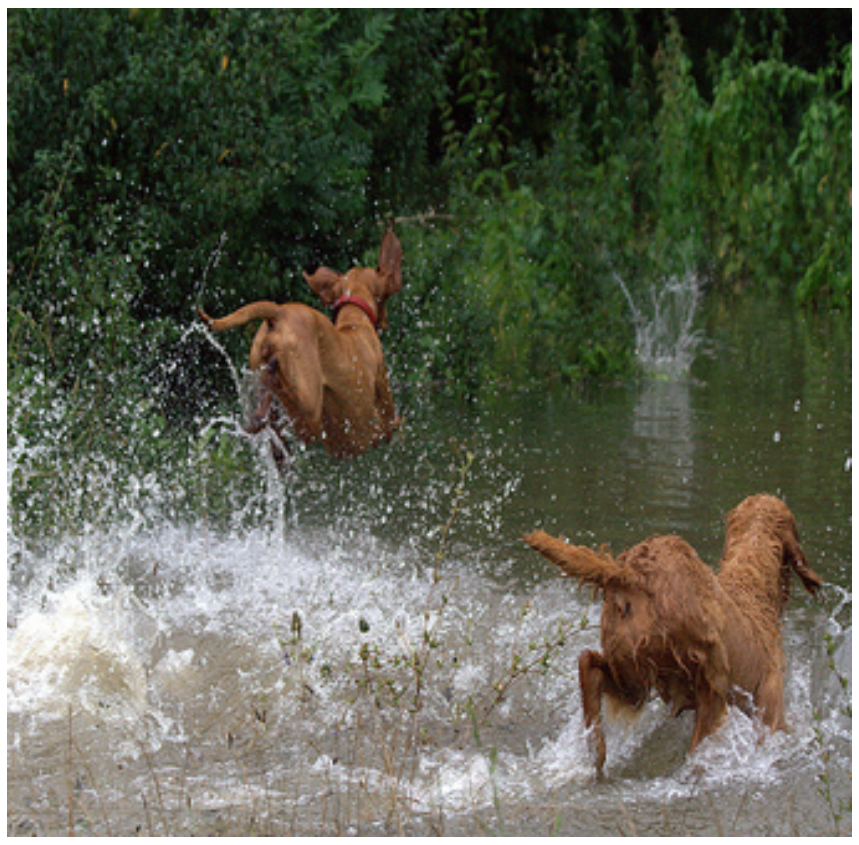

**Crowdsourced Hindi Captions**
दो भूरे कुत्ते जलाशय में दौड़ रहे हैं
दो कुत्ते पानी में जोर से दौड़े जा रहे है
दो कुत्ते पानी में दौड़ लगा रहे है
कुत्ते पानी में दौड़ रहे है
कुत्ते पानी में रेस लगा रहे है

**English Captions from Flickr8k**
Two brown dogs are creating large splashes as they run in a river .
Two brown dogs in the water .
Two brown dogs running through water .
Two brown dogs runs through the water .
Two dogs splash through the water .

**Translated Hindi Captions (for reference)**
Two brown dogs running in the reservoir
Two dogs running fiercely in the water
Two dogs are running in the water
Dogs running in water
Dogs are racing in water

Figure 8: An Example Image and the Corresponding English (part of the dataset) and Hindi (generated by the workers using only the image) captions.